\pdfoutput=1

\documentclass[11pt]{article}

\usepackage[]{ACL2023}

\usepackage{times}
\usepackage{latexsym}
\usepackage{xspace}
\usepackage{graphicx}
\usepackage{bookmark}
\usepackage[all]{nowidow}

\newif\ifcomment\commenttrue

\usepackage[a-1b]{pdfx}

\usepackage{framed}
\usepackage{mdwlist}
\usepackage{siunitx}
\usepackage{latexsym}
\usepackage{colortbl}
\usepackage{nicefrac}
\usepackage{booktabs}
\usepackage{fnpct}
\usepackage{amsfonts}
\usepackage[T1]{fontenc}
\usepackage{bold-extra}
\usepackage{amsmath}
\usepackage{amssymb}
\usepackage{bm}
\usepackage{graphicx}
\usepackage{mathtools}
\usepackage{microtype}
\usepackage{multirow}
\usepackage{multicol}
\usepackage{xpatch}
\usepackage{bbm}

\usepackage{latexsym,comment}
\usepackage[normalem]{ulem}

\newcommand*{\missingreference}{{\Huge \colorbox{red}{?reference?}}}
\newcommand*{\missingcitation}{{\Huge \colorbox{red}{?citation?}}}

\makeatletter
\xpatchcmd{\@setref}{\bfseries}{\missingreference}{}{}
\def\@citex[#1]#2{\leavevmode
    \let\@citea\@empty
    \@cite{\@for\@citeb:=#2\do
        {\@citea\def\@citea{,\penalty\@m\ }%
            \edef\@citeb{\expandafter\@firstofone\@citeb\@empty}%
            \if@filesw\immediate\write\@auxout{\string\citation{\@citeb}}\fi
            \@ifundefined{b@\@citeb}{\hbox{\reset@font\missingcitation}%
                \G@refundefinedtrue
                \@latex@warning
                {Citation `\@citeb' on page \thepage \space undefined}}%
            {\@cite@ofmt{\csname b@\@citeb\endcsname}}}}{#1}}
\makeatother

\newcommand{\gem}[1]{\mbox{\textsc{gem}}}
\newcommand{\abr}[1]{\textsc{#1}}

\newcommand{\hidetext}[1]{}
\newcommand{\ignore}[1]{}

\ifcomment
    \newcommand{\pinaforecomment}[3]{\colorbox{#1}{\parbox{.8\linewidth}{#2: #3}}}

    \newcommand{\prtodo}[1]{\pinaforecomment{lightblue}{pr}{#1}}
    \newcommand{\prtodoi}[1]{\pinaforecomment{lightblue}{pr}{#1}}
\else
    \newcommand{\pinaforecomment}[3]{}
    \newcommand{\prtodo}[1]{}
    \newcommand{\prtodoi}[1]{}
\fi

\newcommand{\smallurl}[1]{ \begin{tiny}\url{#1}\end{tiny}}

\definecolor{lightblue}{HTML}{3cc7ea}
\definecolor{CUgold}{HTML}{CFB87C}
\definecolor{grey}{rgb}{0.95,0.95,0.95}
\definecolor{ceil}{rgb}{0.57, 0.63, 0.81}
\definecolor{UMDred}{HTML}{ed1c24}
\definecolor{UMDyellow}{HTML}{ffc20e}

\usepackage[T1]{fontenc}

\usepackage[utf8]{inputenc}

\usepackage{booktabs}

\usepackage{microtype}

\usepackage{inconsolata}

\definecolor{russia}{RGB}{117, 125, 145}
\definecolor{austria}{RGB}{255, 0, 0}

\definecolor{turkey}{RGB}{185, 166, 28}
\definecolor{italy}{RGB}{34, 139, 34}
\definecolor{germany}{RGB}{160, 138, 117}

\newcommand{\fullname}{\textbf{P}ersonalized \textbf{H}elp for \textbf{O}ptimizing \textbf{L}ow-Skilled \textbf{U}sers' \textbf{S}trategy}

\newcommand{\name}{\abr{pholus}\xspace}

\newcommand{\cicero}
{\textsc{cicero}\xspace}

\title{Personalized Help for Optimizing Low-Skilled Users' Strategy}

\author{Feng Gu$^{1}$ \hspace{0.5cm} Wichayaporn Wongkamjan$^{1}$ \textbf{Jonathan K. Kummerfeld}$^{2}$  \\   \hspace{0.5cm} \textbf{Denis Peskoff}$^{3}$ \hspace{0.5cm} \textbf{Jonathan May}$^{4}$ \hspace{0.5cm} \textbf{Jordan Lee Boyd-Graber}$^{1}$ \\
  $^{1}$University of Maryland \hspace{0.5cm}
    $^{2}$University of Sydney \\
  $^{3}$Princeton University \\
  $^{4}$Information Sciences Institute, University of Southern California \\
  \texttt{\{fgu1, wwongkam\}@umd.edu} \hspace{0.5cm} \texttt{jbg@.umiacs.umd.edu}
}

\begin{document}
\maketitle
\begin{abstract}
\abr{AI}s can beat humans in game environments; however, how helpful those agents are to humans remains understudied.
We augment \cicero, a natural language agent with superhuman performance in \textit{Diplomacy}, to generate both move and message advice based on player intentions.
In a dozen \textit{Diplomacy} games, novice and experienced players, with varying advice settings, benefit from some of the generated advice. Advice helps novices compete with experienced players and in some instances even surpass them. Just reading advice can be advantageous, even if players do not follow it.\footnote{Code available at \url{https://github.com/Obertura777/pholus}}
\end{abstract}












\section{Leveraging Human-\abr{ai} Collaboration}


%
%






\abr{AI} and humans are frequent collaborators: in writing~\cite{10.1145/3491102.3502030}, making decisions~\cite{Bansal2019BeyondAT}, and creating artwork~\cite{10.1145/3490099.3511135}.
The most fruitful collaborations are those in which humans and computers have complementary skills, such as \abr{ai} analyzing medical imaging to identify anomalies and doctors interpreting these findings.
%
%
We posit that the board game \textit{Diplomacy} is an apt testbed for studying this type of collaboration.
\citet{wongkamjan-etal-2024-victories} study \cicero{}~\cite{cicero}, the best \textit{Diplomacy}-playing \abr{ai} capable of communicating in natural language, and show that while the state-of-the-art \abr{ai}s have near-optimal move strategy, human players remain better at communication.
%
\begin{figure}
    \centering
    \includegraphics[width=0.45\textwidth]{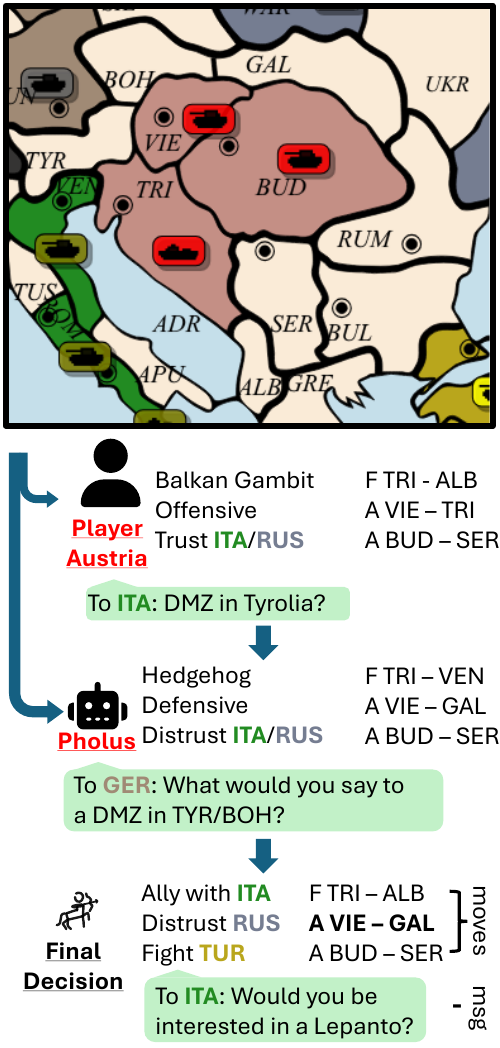}
    \caption{\name{} generates move and message advice based on the game state and the player's past messages. Initially, as \textcolor{austria}{Austria}, the player considers the Balkan Gambit, assuming cooperation from \textcolor{italy}{Italy} and \textcolor{russia}{Russia} to capture Serbia and Greece. \name{} suggests the Hedgehog, a more defensive opening. The player eventually adopts a synthesized strategy: forming an anti-\textcolor{turkey}{Turkey} alliance with \textcolor{italy}{Italy} (Lepanto) while using the Vienna unit to defend against a potential \textcolor{russia}{Russian} attack in Galicia. The final decision highlights altered moves.}
    \label{fig:intro_example}
\end{figure}




%



We introduce \fullname{} (\textbf{\name{}}),\footnote{We use the name \name{} because he was a centaur, a mythological combination of a human and a horse. After Gary Kasparov's defeat to Deep Blue~\cite{wilkenfeld-19}, he advocated for ``\emph{centaur chess}''---where humans and computers play together---as a way of maintaining competitive games.} a natural language agent that provides both moves and messages generated by \cicero{} as advice to \textit{Diplomacy} players in real-time. The core distinction between them is that \cicero{} is a game-playing agent whereas \name{} is an advisor that does not actively participate in the game. Players' moves and message history influence \name{}'s advice.

We run a user study and collect a dataset with twelve games, $1{,}070$ player turns, and $117$ playing hours. 
\name{} enables novices---who barely know the rules of \textit{Diplomacy}---to compete with experts (Figure~\ref{fig:advice_reg}). But this does not just mean the novices blindly follow the advice.
First, they use \name{}'s strategic insights to inform their communication strategies with other players.
Second, \name{} helps experienced players, although they are less inclined to take the advice than the novices. Overall, both advice types from \name{} improve players' game outcomes (Section~\ref{sec: quantitative}). Our research enables human-\abr{ai} collaboration and offers valuable insights into the potential of using \abr{ai} to enhance human learning experiences.

On a broader scale, our study explores the potential for \abr{ai}s like \name{} to enhance learning in unfamiliar environments. \abr{ai} agents surpass traditional rule-based methods by offering more flexible and personalized learning experiences. Integrating tailored guidance into human intelligence, these systems provide unique learning experiences for inexperienced individuals. Future research directions in human-\abr{ai} collaboration include generating advice based on high-level intentions and goals, reducing over-reliance on skilled \abr{ai}s, and facilitating learning processes.









\section{\textit{Diplomacy} as a Cooperative Testbed}


%


\textit{Diplomacy} is a seven-player, turn-based board game.
The goal is to obtain more than half of the board's possible points.\footnote{Represented by a subset of spaces / territories on the map termed \textit{supply centers}.}
Critically, turns are simultaneous, with moves written in secret by players and then revealed.
This means that players must communicate to collaborate effectively.

\subsection{Experiment Setup}
We recruit \textit{Diplomacy} players online. For experienced players, we advertise in the \textit{Diplomacy} community (specifically players active on \textit{webDiplomacy} and \textit{Backstabbr}, as well as in-person tournament attendees). To find novice players, we contact board game enthusiasts in university clubs. A novice player is someone who has no prior \textit{Diplomacy} experience and is unfamiliar with its rules.


We modify a game engine and interface~\cite{NEURIPS2019_84b20b1f} and maintain the same game format used by \citet{wongkamjan-etal-2024-victories}. Each game involves two to five human players. Games last about three hours, with each turn taking ten minutes. 

As illustrated in Figure~\ref{fig:intro_example}, \name{} passively observes the game. If \cicero{} is an active participant, it would have submitted moves and sent messages based on the game state and its message history. Instead, \name{} presents these moves and messages as advice to players. Each time the player sends a message, \name{} recomputes advice given the new context and presents it to the user.
Every turn, we randomly assign each player to one of the following settings: \\
\noindent \textbf{1) No advice}: \name{} does not offer any information, meaning the player receives no assistance from \name{}. \\
\noindent \textbf{2) Message advice}: \name{} suggests \textit{to whom} a player should send a message and \textit{what} the message content could be. \\
\noindent \textbf{3) Move advice}: \name{} recommends a set of moves (or unit orders) to the player.  \\
\noindent \textbf{4) Message and move advice}: Combines the previous two types.

In total, we collect data from twelve games involving fourty-one players.
This includes over $3{,}600$ entries of move advice and $4{,}300$ pieces of message advice (Table~\ref{tab:advice}).

\begin{table}


\centering
\resizebox{\columnwidth}{!}{
\begin{tabular}{lrrrr}
 \hline
\textbf{} & \multicolumn{2}{c}{\textbf{Move Advice}} & \multicolumn{2}{c}{\textbf{Message Advice}}\\
&Accepted&Total&Accepted&Total\\
 \hline
 \textbf{Novices}&32.6\%&872&6.3\%&1413\\
\textbf{Veterans}&6.4\%&2807&3.4\%&2912\\
 \hline
\end{tabular}
}
\caption{Statistics of advice generated by \name{} and accepted by players. \textit{Diplomacy} novices are more willing to accept move and message advice than veterans. Move advice is more frequently accepted than message advice for both novices and veterans.}
\label{tab:advice}
\end{table}

\subsection{Evaluation Metrics}
To assess the effectiveness of \name{}'s advice, we consider the net gain or loss of points in each turn as the effect of advice. We train a linear regression model with regularization to examine the advice's effectiveness. The model includes features such as which of the seven Great Powers is assigned to the player, the number of turns that have passed, the player's type (novice or veteran), and the advice setting. We encode the Power, player type, and advice setting as one-hot vectors.

To evaluate players' reliance on \name{}, we use both qualitative and quantitative methods. In addition to computing move advice acceptance frequency, we also measure agreement and equivalence between the move suggested by \name{} and a player's moves. Agreement is the proportion of moves that appeared in both the players' move set and \name{}'s advice set in a given turn. The sets are equivalent if they overlap entirely. Formally, we define move agreement $\mathcal{A}$ in turn $i$ as $\mathcal{A}_{x_i, y_i} = |x_i \cap y_i| / |x_i|$ and equivalence $\mathcal{E}$ as $\mathcal{E}_{x_i, y_i} = 1$ if $ x_i = y_i$ and $\mathcal{E}_{x_i, y_i} = 0$ otherwise, 
where \( x_i \) is the player's move set and \( y_i \) is \name{}'s move advice set in turn \( i \)\footnote{For any $i$, \(|x_i| = |y_i|\).}. Agreement is particularly useful for capturing the overlap when players reject the complete move advice set but follow individual advice from \name{}.

\section{\textsc{Pholus} Provides Helpful Advice}
\subsection{Quantitative Analysis}
\label{sec: quantitative}

\textbf{Non-advice factors parallel previous findings.} Playing as France offers the most strategic advantage~\cite{dipstats}. \cicero{} playing as Germany or Italy is correlated with better game outcomes, while playing as Austria, England, or Turkey is correlated with worse game outcomes~\cite{wongkamjan-etal-2024-victories}. Additionally, \cicero{} dominates: of twelve games, \cicero{} won eight. 

\begin{figure}
    \centering
    \includegraphics[width=0.45\textwidth]{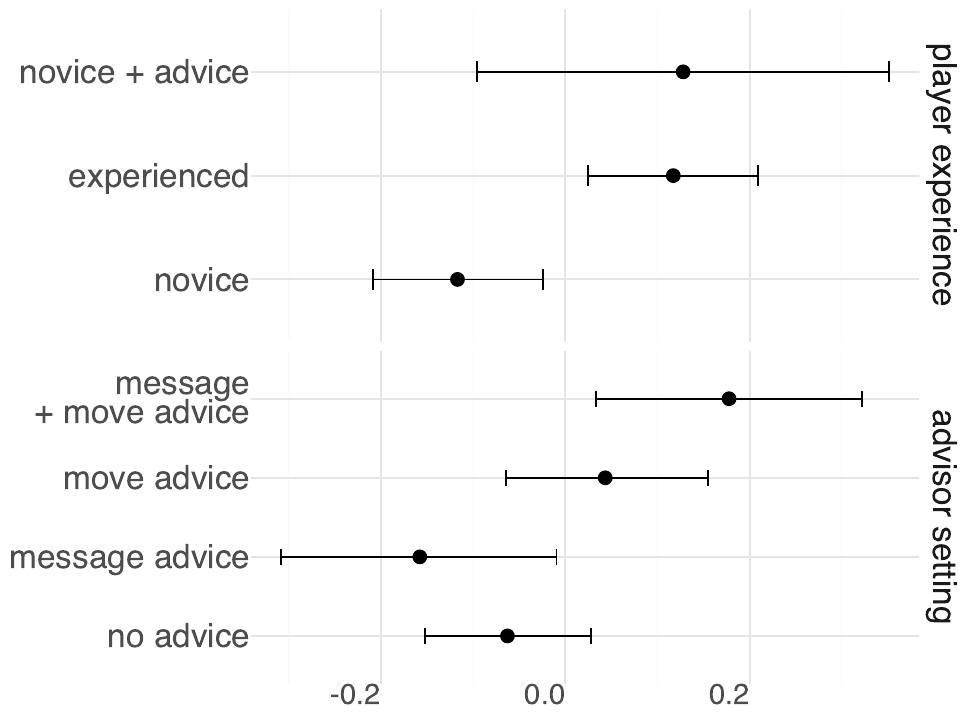}
    \caption{Regression coefficients for advice settings and player skills to predict supply center gains. Not receiving any advice from \name{} is slightly disadvantageous. Move advice has a positive correlation with player performance. Receiving both forms of advice has the greatest positive impact. As expected, not having previous exposure to \textit{Diplomacy} is indicative of bad performance. However, with the help of \name{}'s advice, \textit{Diplomacy} novices are on the same level as veterans and have the potential to defeat experienced players.}
    \label{fig:advice_reg}
\end{figure}

\textbf{Advice helps.} Playing a game without advice puts players at a disadvantage.  The feature associated with no advice has a negative coefficient of approximately $-0{.}05$ (Figure~\ref{fig:advice_reg}).
The coefficients suggest a slight positive correlation between receiving move advice and point gains. Players who receive both move and message advice gain more points than those who receive only move advice. Interestingly, only having message advice negatively affect players' game outcomes.


\textbf{Novices can outperform experienced players with the help of \name{}.} Players with no prior experience in \textit{Diplomacy} naturally face a disadvantage against seasoned players. This often results in novices being eliminated relatively early in the game. Even if they remain in the game, losing supply centers is almost inevitable.
However, novice players receiving advice play better: in five games where novices received message and move advice, only one player
was eliminated before the game concluded (typically $3$--$4$ players in a game are eliminated). In the other four games, novices ended the game with more supply centers than they started with.

\textbf{Novices are more likely to follow \name{}'s advice.} Experienced players tend to disregard advice. They accept only $3.4\%$ of message advice and $6.4\%$ of move advice from \name{}. Although novice players are also hesitant to accept message advice, doing so $6.3\%$ of the time, this rate is nearly double that of experienced players. Novice players follow move advice approximately one-third of the time, with an acceptance rate of $32.6\%$. Both novice and experienced players tend to take more move advice than message advice. 

\textbf{Novices do not fully trust move advice from \name{}.} Across all games, \name{} generates $333$ instances of individual move advice for novices, organized into $134$ sets. At the start of turns, average move agreement is $80\%$ and average equivalence is $46\%$, indicating strong alignment between novices' initial idea for moves and \name{}'s move advice. However, by the end of each turn, the average agreement drops by $10\%$ and the average equivalence decreases by $8\%$, indicating that novice players do not follow the move advice blindly.

\subsection{Qualitative Analysis}
\label{sec:qualitative}
While we can compute equivalence $\mathcal{E}$ for moves, this is more difficult for messages. To better understand why players reject \name{}'s advice more than they accept it (Table~\ref{tab:advice}), we  qualitatively investigate the differences between \name{}'s suggested messages and the actual messages sent by players. 
To analyze message content, we use Abstract Meaning Representation~\citep[\abr{amr},][]{amr-banarescu-etal-2013-abstract} to extract \textit{Diplomacy}-specific tokens. We parse player messages and the corresponding message advice from \name{} to \abr{amr}. We then measure the similarity of the two parses using \abr{smatch} score~\cite{cai-knight-2013-smatch}.
Many pairs have high \abr{smatch}, indicating that players often incorporate parts of \name{}'s advice into their messages. For example,
\name{} suggests ``\emph{bounce in Galicia again?}'' while the player wrote
``\emph{Do you want to bounce in Galicia again?}''
Despite being written differently, these clearly have the same meaning, and indeed,  \abr{smatch} gives the pair a score of $0.74$.

We also notice message-advice pairs with low \abr{smatch} scores, where human players have different objectives in mind. For instance, in the fifth game, Italy captures Warsaw from Russia and anticipates losing it in the next turn due to Russia's stronger nearby presence. When Russia inquires about the unexpected attack, \name{} suggests using the fallacy of deflection to feign ignorance: ``\emph{Turkey has been the only one to heed my concerns, despite my reservations,}'' and, ``\emph{I thought you were lying}''.
However, the player disregards the advice of talking to Russia. Instead, the player seeks help from Turkey, who has an adjacent unit, to ``\emph{support Warsaw's hold}''. The player then secures the support from Turkey and successfully keeps Warsaw from subsequent Russian attack. These pairs yield \abr{smatch} scores of $0$.

Our analysis indicates that \abr{smatch} scores match with our intuitions about textual similarity.
Given the many high \abr{smatch} scores, we can conclude that many messages that are sent by players are minor variations on the provided advice. We provide more examples in Appendix~\ref{sec:smatch_ex}.
For additional qualitative insights, we survey players on the effectiveness of the advice. We summarize the survey results in Appendix~\ref{sec:sr}.





\section{Related Work}
\textbf{Appropriate Reliance on \abr{ai}:} The topic of human reliance on \abr{ai} is central to current research in machine learning and explainable \abr{ai}. Prior work measures reliance in \abr{ai}-assisted decision making~\cite{10.1145/3581641.3584066, 10.1145/3610219, 10.1145/3613904.3642621,  zhou2024relaiinteractioncenteredapproachmeasuring}, and explores reducing over-reliance~\cite{10.1145/3449287, schemmer2022ifollowaibasedadvice, 10.1145/3579605}. Some researchers have examined how explanations affect human reliance on \abr{ai}~\cite{starke2021fairnessperceptionsalgorithmicdecisionmaking, 10.1145/3476068}. However, empirical evidence from multiple domains shows conflicting results: while some show that \abr{ai} explanations improve human decision making, others find evidence of over-reliance on \abr{ai} explanations even when they are incorrect~\cite{10.1145/3287560.3287590,  10.1145/3377325.3377498,  10.1145/3351095.3372852,  10.1145/3397481.3450650, 10.1145/3411764.3445717, 10.1145/3411764.3445315, 10.1145/3479552,kim2022hiveevaluatinghumaninterpretability, chenglei}. For \name{}, humans remain relatively conservative toward \abr{ai} advice. Even novice \textit{Diplomacy} players do not blindly follow the advice.

\textbf{\abr{AI} as Player Companion:} \abr{AI} agents have a long history of superhuman gameplay. In 1996, \abr{IBM}'s Deep Blue defeated the reigning world chess champion, Garry Kasparov, although it lost several other games in the same match~\cite{deepblue}. More recently, DeepMind's AlphaGo~\cite{alphago} consistently defeated top-rated Go players, a game with exponentially complex computational space, and later changed professional Go players' play style. Multi-agent reinforcement learning systems like AlphaStar~\cite{alphastar} and OpenAI Five~\cite{five} also show high performance in computer games through self-play.

However, these experiments focus only on game outcomes rather than how they can shape human gameplay. Some studies on \abr{nlp} communicative agents aim to generate guidance in a grounded environment~\cite{10.1145/1822348.1822365}. \citet{Tremblay2013AdaptiveCI} develop an adaptive \abr{ai} companion that adjusts its behavior based on the player's experience. \citet{doi:10.1177/00187208231190459} assess human reliance on \abr{ai}-based advice by examining the skill level of \abr{ai} agents and the presentation of advice. While these studies show that \abr{ai}s outperform non-adaptive agents in guiding players, they do not consider player intention when generating guidance. In comparison, \name{} takes players' past messages and moves entered when generating personalized advice.

\textbf{Augmented Learning:} This is an educational approach that enhances and personalizes the learning experience. Traditionally, peer interaction simulates social interaction and helps learning~\cite{pedagents}. Recent advancements in \abr{ai} and \abr{nlp} agents, suggest adaptive pedagogical interactions between humans and these agents to help learning in new environments~\cite{esial, pmtpslive, doi:10.1177/1529100615569721, https://doi.org/10.1609/aimag.v39i2.2793}. \citet{Pei-et-al-2023-dnd-theory-of-mind} apply the theory of mind to generate guidance for players in \textit{Dungeons and Dragons}. \citet{10.1145/3392063.3394400} develop a narrative-based tutoring system and show that it helps effective learning for children. In this study, we apply the concept of augmented learning to help novices understand the game of \textit{Diplomacy}.

\section{Conclusion}

Human-\abr{ai} collaboration depends on a range of factors.
Using the board game \textit{Diplomacy}, \name{} provides real-time move and message advice tailored to intentions of both novice and experienced players.
Surprisingly, even though only some advice is accepted, it can have a substantial impact on outcomes, particularly for novice players.
This is because advice can positively inform choices even if the advice isn't strictly followed.
Our experiments enable further study of human-\abr{ai} collaboration, including modeling explicit intentions and how to better use knowledge within these models.
On a broader scope, future research should consider how \abr{ai} can inform people without making choices for them and measure that impact.

\section{Limitations}
While we can effectively use \name{} to generate both message and move advice for players,
this advice can be too general or may not align with player intentions at times. For example, when a player expresses interest in an alliance with another player, \name{} may give aggressive move advice deemed hostile toward that Power. We suspect that the advice may be optimized more for \cicero's intentions, which come from optimal moves in the supervised training data. Consequently, players who are willing to sacrifice individual optimality for mutual gains may find the advice less useful.

Furthermore, \name{} cannot generate advice based on high-level player intentions. Specifically, \name{} generates move advice based on optimal utility and message advice by inferring intentions from player-input moves. Potential improvements include 1) explaining meta-level intentions (e.g., ally with Germany and prioritize defeating Austria) from player input, and 2) generating targeted move and message advice based on meta-level intentions.

Finally, \name{} is a resource-intensive advisor that runs on high-end \abr{gpu}s that require a large amount of on-chip memory (over 35GB). We use Nvidia's A100 for running \name{}. This limits accessibility for \textit{Diplomacy} players and researchers to efficiently utilize \name{}. The community would benefit from a distilled version of \name{} by reducing computational limits and future adaptations.

\section{Ethical Considerations}
We recruit players individually via email and assign pseudonyms to ensure anonymity, even if players know each other outside the experiment. We adhere to human subject research regulations and the study was approved by our institution's ethics review board (IRBNet ID: 1740681, University of Maryland). We report the experimental procedure in Appendix~\ref{sec:proc} and compensation details in\ref{sec:comp}.

\section*{Acknowledgments}
We thank Meta for open sourcing \cicero{}. We thank the \textit{Diplomacy} community for taking interest in our study. Specifically, our thank goes to Matthew Totonchy, Dr. Abhishek Singhal, Antonio Imperato, Sophia Wiste, and other members of the community who took the time to play against \cicero{}. In addition, we thank Yanze Wang and Sadra Sabouri from University of Southern California for their helpful feedback. Denis Peskoff is supported by the National Science Foundation under Grant No. 2127309 to the Computing Research Association for the CIFellows 2021 Project. This research is supported by the U.S. Defense Advanced Research Projects Agency (DARPA) Other Transaction Award HR00112490374 from the Friction for Accountability in Conversational Transactions (FACT) program. Any opinions, findings, conclusions, or recommendations expressed here are those of the authors and do not necessarily reflect the view of our sponsors.

\bibliography{bib/2024_arr_chiron-advisor, bib/jbg}
\bibliographystyle{acl_natbib}

\appendix

\section{Appendix}
\subsection{Diplomacy}
\textit{Diplomacy} is a board game that has two core components: strategy and communication. Strategic reasoning plays a crucial role in determining the game's outcome, as players' moves directly impact the board's status. Meanwhile, negotiation and deception significantly influence player strategies. Successful cooperation can remove a common adversary from the board, while a well-timed betrayal by a trusted ally can be catastrophic, greatly reducing the chances of winning. Excelling in \textit{Diplomacy} requires not only a thorough understanding of the game's mechanics but also strong communication skills. Consequently, \textit{Diplomacy} is an ideal testbed for studying human-\abr{ai} interaction and appropriate reliance in a grounded environment where outcomes are clearly observable.

Early efforts to develop agents for \textit{Diplomacy} concentrated solely on creating rule-based agents that relied heavily on feature engineering~\cite{Albert}.
These agents only submit moves and are not capable of communication. In 2002, a group of programmers released a communication protocol, \textit{Diplomacy Artificial Intelligence Development Environment} ~\citep[\abr{daide},][]{daide}. \abr{DAIDE} defines a language syntax that enables agents to diplomatically negotiate and describe game actions. Following \abr{daide}, researchers built communicative agents, including Albert~\cite{Albert}, SillyNegoBot~\cite{6078245}, DipBlue~\cite{DipBlue}.

Starting with DipNet~\cite{NEURIPS2019_84b20b1f}, neural networks were applied to the game, leading to the first agents that were competitive with people.
Subsequent studies incorporated reinforcement learning to achieve super-human performance~\cite{gray2021humanlevel, bakhtin2021nopress, 10.5555/3495724.3497234, pmlr-v162-jacob22a}.

\subsection{All Regression Coefficients}
\begin{figure}
    \centering
    \includegraphics[width=0.45\textwidth]{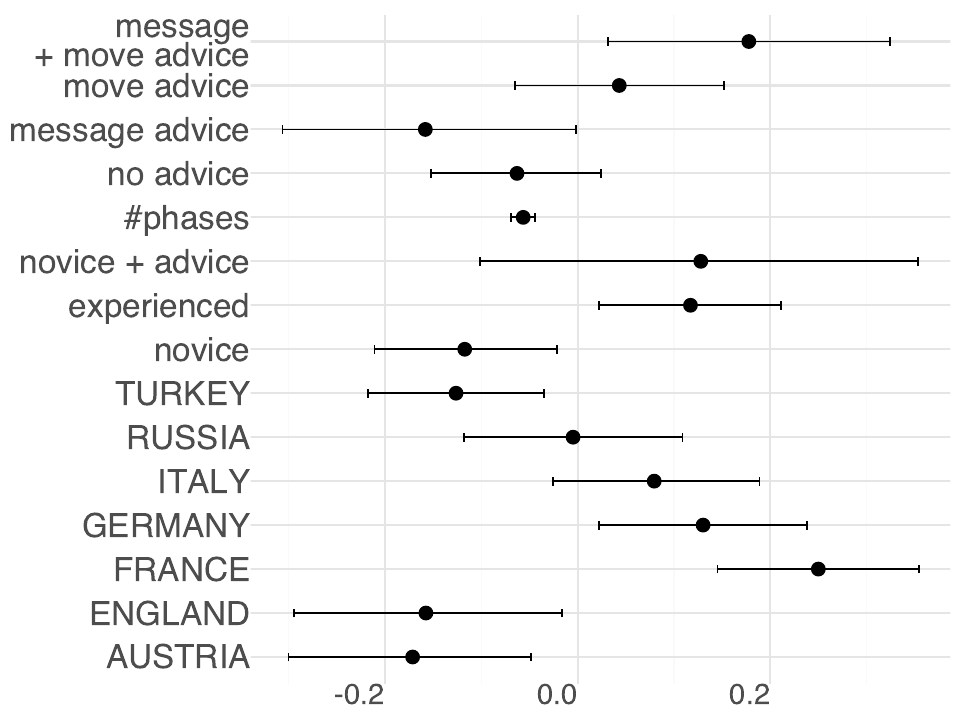}
    \caption{Regression coefficients for all features.}
    \label{fig:all_coefs}
\end{figure}

In Figure~\ref{fig:advice_reg}, we only show regression coefficients related to advice setting and player experience. Figure~\ref{fig:all_coefs} contains coefficients for all regression features. The official \textit{Diplomacy} rule states that supply center control changes only on even turns. However, we consider moving a unit to a center on an odd turn as gaining it, since the unit typically remains there in the next turn.

\subsection{Experiment Procedures}
\label{sec:proc}
The procedures involve playing \textit{Diplomacy} with other participants as
well as \abr{ai} agents and providing feedback throughout the game on
1) the truth value of players' messages 2) the predicted truth values of
theirr opponents’ messages 3) the perceived friendliness of opponents 4) if players think they have been lied to in the previous phase.
Players are asked to participate by placing their game moves in each turn before communicating. Some players in some turns received strategic moves and message advice from \name{}.
Games end after  3 hours or less. After each game, participants are
 asked to fill out a survey asking about their
experience and the \abr{ai} advice, as well as how \abr{ai} agents
performed during the game. 

\subsection{Participant Compensation}
\label{sec:comp}
Players received a \$70 gift card of their choice for each game. In
addition, up to \$29 was rewarded based on performance in the
game.

\subsection{Survey Questions}
We use Google Forms to conduct player surveys. Table~\ref{tab:sq} shows the survey questions.
\begin{table*}
\centering
\begin{tabular}{ll}
\toprule
\textbf{Question} & \textbf{Type}\\ \midrule
I am really good at Diplomacy. & Likert scale \\
How many Diplomacy games have you played before? & Free text\\
I am able to identify all bots. & Likert scale \\
I enjoy talking with the AIs. & Likert scale \\
I was able to make plans with other players in the game.& Likert scale \\
I was able to make plans with the AIs in the game.& Likert scale \\
human players communicated transparently.& Likert scale \\
AI players communicated transparently.& Likert scale \\
Do you have a link to your diplomacy stats?& Free text \\
How helpful is the move advice?& Likert scale \\
Thoughts on the move advice?& Free text \\
How helpful is the message advice?& Likert scale \\
Thoughts on the message advice?& Free text \\
Feel free to let us know your other post-game thoughts.& Free text \\
 
 \\ \bottomrule
\end{tabular}
\caption{List of questions in the survey.}
\label{tab:sq}
\end{table*}

\subsection{\textsc{Pholus}'s message advice and human messages with \abr{smatch} scores}
\label{sec:smatch_ex}
We provide additional examples of cases where human players reject \name{}'s message advice, together with \abr{smatch}. We show some advice (Figure~\ref{fig:rus_high_s} and \ref{fig:ger_high_s}) that experienced players mostly agree with, however, they partially edit to make it more aligned with their goals. We also show advice that is not aligned with the player's goals, for example in Figure~\ref{fig:ita_low_s}.

\begin{figure}[h]
    \small{\texttt{\hspace*{4 mm}Sender: Italy (veteran)\\
    \hspace*{4 mm}Recipient: England\\
    \hspace*{4 mm}Message Advice: absolutely. Have you and Germany and France decided on a plan? That will influence my opening quite a bit.\\
    \hspace*{4 mm}Human Message: absolutely. Have you and Germany and France decided on a plan? That will influence my opening quite a bit.\\
    \hspace*{4 mm}\abr{smatch}: 1.0}}
    \caption{Human uses \name{}'s advice directly without modification.}
\end{figure}

\begin{figure}[h]
    \small{\texttt{\hspace*{4 mm}Sender: Italy (novice)\\
    \hspace*{4 mm}Recipient: Turkey\\
    \hspace*{4 mm}Message Advice: hey can you move to Con to help the game?\\
    \hspace*{4 mm}Human Message: Hey Turkey, could you move to CON? Thanks!\\
    \hspace*{4 mm}\abr{smatch}: 0.7895}}
    \caption{Italy paraphrases the advice with the same underlying intention.}
\end{figure}

\begin{figure}[h]
    \small{\texttt{\hspace*{4 mm}Sender: England (veteran)\\
    \hspace*{4 mm}Recipient: Russia\\
    \hspace*{4 mm}Message Advice: I might convoy an army over to get it but I won't move into the Barents\\
    \hspace*{4 mm}Human Message: I'm convoying an army into Norway, but it's sincerely to preserve navies for the fight that France clearly wants\\
    \hspace*{4 mm}\abr{smatch}: 0.5833}}
    \caption{\label{fig:rus_high_s}The advice and human message share a general intent (reassure Russia), but differ slightly.}
\end{figure}

\begin{figure}[h]
    \small{\texttt{\hspace*{4 mm}Sender: Germany (veteran)\\
    \hspace*{4 mm}Recipient: Italy\\
    \hspace*{4 mm}Message Advice: I can't, gotta keep France out of Bur\\
    \hspace*{4 mm}Human Message: I'm sorry I need that army to bounce in Bur. I can help next turn though\\
    \hspace*{4 mm}\abr{smatch}: 0.5641}}
    \caption{\label{fig:ger_high_s}The advice and human message point out the same goal that rejects the other player's proposal, instead using their unit to bounce in Bur.}
\end{figure}

\begin{figure}[h]
    \small{\texttt{\hspace*{4 mm}Sender: Italy (novice)\\
    \hspace*{4 mm}Recipient: Turkey\\
    \hspace*{4 mm}Message Advice: Russia is going to attack you!\\
    \hspace*{4 mm}Human Message:  I'm going to hold Mars\\
    \hspace*{4 mm}\abr{smatch}: 0.5532}}
    \caption{\label{fig:ita_low_s}\name{} wants to warn Turkey regarding Russia's attack, however, the player disregards that and instead, shares their plan.}
\end{figure}

\begin{figure}[h]
    \small{\texttt{\hspace*{4 mm}Sender: Italy (novice)\\
    \hspace*{4 mm}Recipient: England\\
    \hspace*{4 mm}Message Advice: Are you going to take Belgium?\\
    \hspace*{4 mm}Human Message: Sure, let's get rid of france\\
    \hspace*{4 mm}\abr{smatch}: 0.4}}
    \caption{An example of low \abr{smatch}. \name{} advises Italy to inquire about a specific game move, but the player discusses a high-level game plan.}
\end{figure}

\subsection{Survey Summarization}
\label{sec:sr}
Many participants find \name{}'s move advice helpful. Experienced players observe that the advice often aligns with their own moves and offers ``\emph{some interesting ideas.}'' However, the advice can be suboptimal and short-sighted in complex scenarios, where it fails to consider interaction between allies. Players find message advice useful for simple, quick communication but inadequate for more complex or specific situations, especially when it does not align with their strategies or alliances. Players mention that the messages include common communication terms, and they ``\emph{regret not using this feature more.}'' However, the advice is less helpful for specific planning and often does not align with player alliances and intentions.

\end{document}